\documentclass[conference,table,xcdraw]{IEEEtran}
\IEEEoverridecommandlockouts
\usepackage{cite}
\usepackage{amsmath,amssymb,amsfonts}
\usepackage{algorithmic}
\usepackage{graphicx}
\usepackage{textcomp}
\usepackage{array,multirow,lscape}
\usepackage[table,xcdraw]{xcolor}
\usepackage{booktabs}
\usepackage{lipsum}
\usepackage{subcaption}
\usepackage{float}
\usepackage{bm}
\usepackage{colortbl} 
\usepackage{soul,color}
\usepackage{hyperref}
\def\BibTeX{{\rm B\kern-.05em{\sc i\kern-.025em b}\kern-.08em
    T\kern-.1667em\lower.7ex\hbox{E}\kern-.125emX}}
\begin{document}

\title{Cross Knowledge Distillation between\\ Artificial and Spiking Neural Networks}

\author{
    \IEEEauthorblockN{
        Shuhan Ye\textsuperscript{2}\IEEEauthorrefmark{2}\thanks{\IEEEauthorrefmark{2} Equal Contribution},
        Yuanbin Qian\textsuperscript{2}\IEEEauthorrefmark{2},
        Chong Wang\textsuperscript{1,2}\IEEEauthorrefmark{1}\thanks{\IEEEauthorrefmark{1} Corresponding author: Chong Wang},
        Sunqi Lin\textsuperscript{2},
        Jiazhen Xu\textsuperscript{2},
        Jiangbo Qian\textsuperscript{1,2},
        and Yuqi Li\textsuperscript{2}
    }
    \IEEEauthorblockA{\textsuperscript{1}Merchants' Guild Economics and Cultural Intelligent Computing Laboratory, Ningbo University, Ningbo, China}
    \IEEEauthorblockA{\textsuperscript{2}Faculty of Electrical Engineering and Computer Science, Ningbo University, Ningbo, China}
}



\maketitle
\begin{abstract}
    Recently, Spiking Neural Networks (SNNs) have demonstrated rich potential in computer vision domain due to their high biological plausibility, event-driven characteristic and energy-saving efficiency. Still, limited annotated event-based datasets and immature SNN architectures result in their performance inferior to that of Artificial Neural Networks (ANNs). To enhance the performance of SNNs on their optimal data format, DVS data, we explore using RGB data and well-performing ANNs to implement knowledge distillation. In this case, solving cross-modality and cross-architecture challenges is necessary. In this paper, we propose cross knowledge distillation (CKD), which not only leverages semantic similarity and sliding replacement to mitigate the cross-modality challenge, but also uses an indirect phased knowledge distillation to mitigate the cross-architecture challenge. We validated our method on main-stream neuromorphic datasets, including N-Caltech101 and CEP-DVS. The experimental results show that our method outperforms current State-of-the-Art methods. The code will be available at \href{https://github.com/ShawnYE618/CKD}{https://github.com/ShawnYE618/CKD}

\end{abstract}

\section{Introduction}
\label{sec:intro}
Spiking Neural Networks (SNNs), regarded as the third generation of neural networks \cite{m:97}, have gained significant attention due to their high biological plausibility, event-driven characteristics \cite{r:19} and energy-saving efficiency\cite{snn}. The ambition of SNNs aligns with the algorithm-hardware co-design paradigm of neuromorphic computing, aiming to serve as a low-power alternative to traditional machine intelligence based on Artificial Neural Networks (ANNs) \cite{r:19}. Unlike ANNs, which use continuous signals and high-power multiply-accumulation operations, SNNs utilize binary spike transmit mechanism, thus leading to substantial improvements in energy efficiency when deployed on neuromorphic hardware \cite{loihi}.

\par However, many aspects of SNN remain underdeveloped. Binary spike activation maps in SNNs have limited information capacity compared to full-precision activation maps in ANNs \cite{guo:23}. As a result, SNNs struggle to retain sufficient information from membrane potentials during the quantization process, leading to information loss and a subsequent accuracy drop. Moreover, the data modalities most suitable for ANNs and SNNs differ significantly; ANNs are well suited for dense and synchronous (RGB) data, whereas SNNs excel with sparse and asynchronous (DVS) data. Capturing DVS data is both time-consuming and costly\cite{ucfcrimedvs}, so available datasets are not only more challenging to collect than RGB frame datasets, but are also typically smaller in scale, which hinders the generalizability of SNN for event-based vision tasks. These challenges put current SNN-based models at a disadvantage compared to their ANN counterparts. Therefore, it is logical to investigate how more effective ANNs, along with RGB datasets that are better suited and more readily available, can be leveraged to enhance the performance of SNNs.
\begin{figure}[t]
    \centering
    \includegraphics[width=\columnwidth]{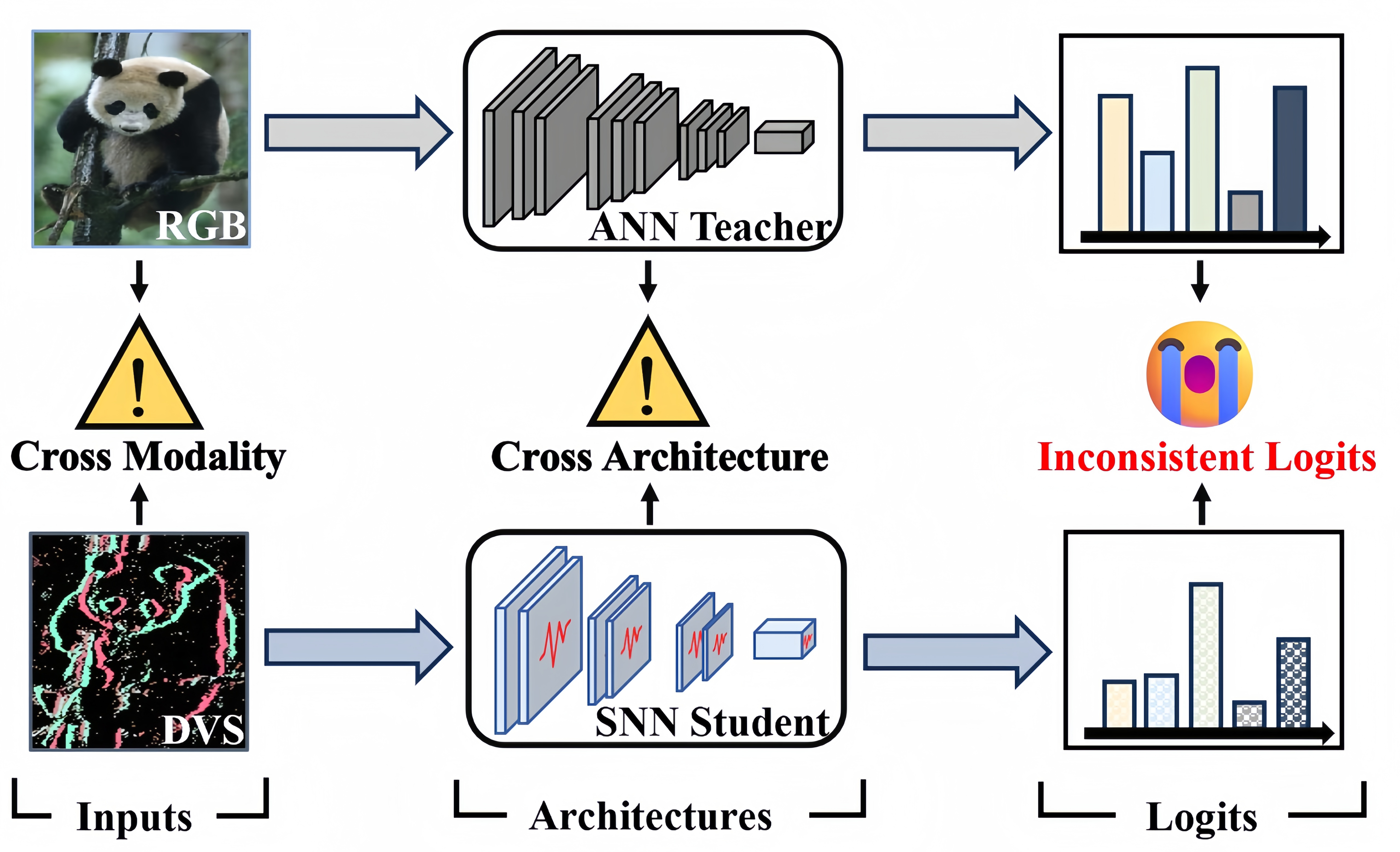}
    \caption{The challenge of cross-modality and cross-architecture in knowledge distillation between ANNs and SNNs.}
    \label{fig:fig0}
    \vspace{-15pt}
\end{figure}

\par Knowledge distillation (KD) \cite{KD} has proved its effectiveness in similar contexts.
However, there is limited work for SNN-based knowledge distillation, the few existing cross-architecture approaches failed to consider the correspondence between modality and architecture\cite{ann2snn1}. In these studies, both models either use RGB data or DVS data. Using ANNs to extract sparse DVS data for distilling SNNs does not fully leverage the potential of ANNs while training SNNs with dense RGB data presents certain risks\cite{respike}.

As a result, ensuring modality-architecture correspondence is crucial. We aim to leverage non-event data and ANNs to mitigate the limitations imposed by the small dataset scale and the limited performance of SNNs. Therefore, employing cross-architecture knowledge distillation is both necessary and challenging. Due to the inherent differences between RGB and DVS data, the distillation process also presents a cross-modality challenge, as shown in Fig. \ref{fig:fig0}.


\par 
Recently, He et al.\cite{he:24} proposed a method which transfers knowledge from RGB data to DVS data in a shared SNN model. Inspired by their work, we present cross knowledge distillation (CKD), an indirect phased distillation method to improve the efficiency of RGB data. CKD utilizes a well-trained ANN model to extract knowledge from RGB data and guide the semantically similar cross-modal data stream within the weight-shared SNN model. Subsequently, superior knowledge is transferred to another dynamic stream via cross-modality knowledge transfer. Through our method, which actually employs a RGB-DVS hybrid stream as an intermediary, the gaps of cross-modality and cross-architecture are effectively bridged. 
Our main contributions are as follows:
\begin{enumerate}
    \item We are the first work to accomplish knowledge distillation that bridges both architecture and modality, simultaneously addressing both RGB data and DVS data, between ANN and SNN.
    \item We leverage the semantic similarity between RGB and DVS data from the same category to achieve modality and architecture correspondence. By employing an RGB-DVS hybrid stream, we enable an indirect phased distillation, allowing ANNs to process the RGB data, while guiding SNNs to achieve maximum performance improvement on the DVS data best suited for them. This also lays the foundation for SNNs to tackle other temporally demanding tasks in the future.
    \item Our sufficient experiments on event-based datasets prove the effectiveness of CKD. Remarkably on N-Caltech101\cite{Ncaltech}, we report a new state-of-the-art top-1 accuracy (97.13\% ), which is 3.68\% higher than our baseline \cite{he:24} (93.45\%), closely approaching the SOTA result of its RGB version Caltech101\cite{Caltech101} on ANN \cite{SotaCaltech101} (98.02\%). Our results demonstrate that with our CKD method, SNNs could perform almost as well as ANNs.
\end{enumerate}


\section{Related Work}
\label{sec:rw}
\noindent \textbf{Spiking Neural Networks.} SNNs draw inspiration from human brain, using discrete spikes for information processing. This method achieves effects comparable to continuous activation functions by accumulating spikes over an additional temporal dimension, making it highly suitable for processing temporal data. Concretely, SNNs replace the traditional activation function by using a spiking neuron model, such as the integrate-and-fire (IF) neuron model and the widely-used Leaky Integrate-and-Fire (LIF) neuron model\cite{LIF}. The LIF neuron model integrates incoming spikes over time, with its membrane potential and spiking behavior governed by the following equations:

\begin{equation}
    \bm{u}^{t+1,l} = \tau \bm{u}^{t,l} + \bm{W}^{l} \bm{s}^{t,l-1}
\end{equation}
\vspace{-2ex}
\begin{equation}
    \bm{s}^{t,l} = H(\bm{u}^{t,l} - V_{\text{th}})
\end{equation}
\vspace{-2ex}
\begin{equation}
    \bm{u}^{t+1,l} = \tau \bm{u}^{t,l} \cdot (1 - \bm{s}^{t,l}) + \bm{W}^{l} \bm{s}^{t+1,l-1}
\end{equation}
where \( \bm{u}^{t,l} \) denotes the membrane potential of neurons in layer \( l \) at time step \( t \), \( \bm{W}^{l} \) represents the weight matrix of layer \( l \), and \( \bm{s}^{t,l} \) corresponds to the binary spikes emitted by neurons. The Heaviside step function \( H \) determines whether a spike is emitted, based on the comparison between \( \bm{u}^{t,l} \) and the threshold \( V_{\text{th}} \). The leaky factor \( \tau \) controls the temporal decay of the membrane potential.

\noindent \textbf{Multimodal Machine Learning.} Multimodal machine learning refers to the integration and processing of information from multiple data sources or modalities, such as images, text, and audio. In machine learning, multimodal models are designed to understand and fuse different types of data to improve performance on complex tasks by leveraging the complementary strengths of each modality. In this work, we use traditional RGB images and DVS (event) data, which is captured by an event camera, also known as a Dynamic Vision Sensor (DVS). DVS excels in high temporal resolution (microsecond-level), low latency, and high dynamic range (\textgreater 120dB). These advantages make them highly compatible with SNNs. Specifically, they asynchronously output a positive or negative event at a pixel location where the brightness change exceeds a certain threshold, which are then integrated into sparse event streams. The output data format is represented as a four-dimensional array $\bm{e}_i$:
\begin{equation}
\bm{e}_i = (t_i, x_i, y_i, p_i)
\end{equation}
where \( t_i \) is the microsecond-level timestamp, \( x_i \) and \( y_i \) are the 2D spatial coordinates, \( p_i \) represents the polarity (1/0 or 1/-1) and \( i \) denotes the index of the \( i \)-th event in the event stream.
Although some recent works are attempting to convert the event stream into other forms of event representation\cite{er1,er2,er3} to fully leverage the characteristics of DVS data, the best-performing methods so far still encapsulate the event stream into event frames within fixed time intervals.

\noindent \textbf{Knowledge Distillation.} Knowledge distillation is a technique in which a well-performing pre-trained model serves as the teacher, transferring its superior knowledge to a smaller student model for accuracy improvement. Recently, there has been increasing research on knowledge distillation for ANNs, including both intra-architecture and cross-architecture distillation \cite{tip, DKD,crossa2}, but this research for SNNs remains relatively limited. This is mainly due to the inherent limitations of the performance of SNNs, where distillation within the same architecture is not very effective \cite{k:20}. Distilling knowledge from ANNs to SNNs faces tricky cross-architecture challenges. Moreover, ANNs are well-suited for traditional dense RGB data, while SNNs excel in processing sparse DVS data. The mismatch between modality and architecture will lead to inferior performance or limited task applicability \cite{respike}. However, previous works have overlooked this important consideration \cite{ann2snn1}. In this work, we exploit the semantic similarity of cross-modal data within the same category to bridge the cross-modality gap, and use an indirect phased distillation method to bridge the cross-architecture gap.

\begin{figure*}[htbp]
    \centering
    \includegraphics[width=1\textwidth]{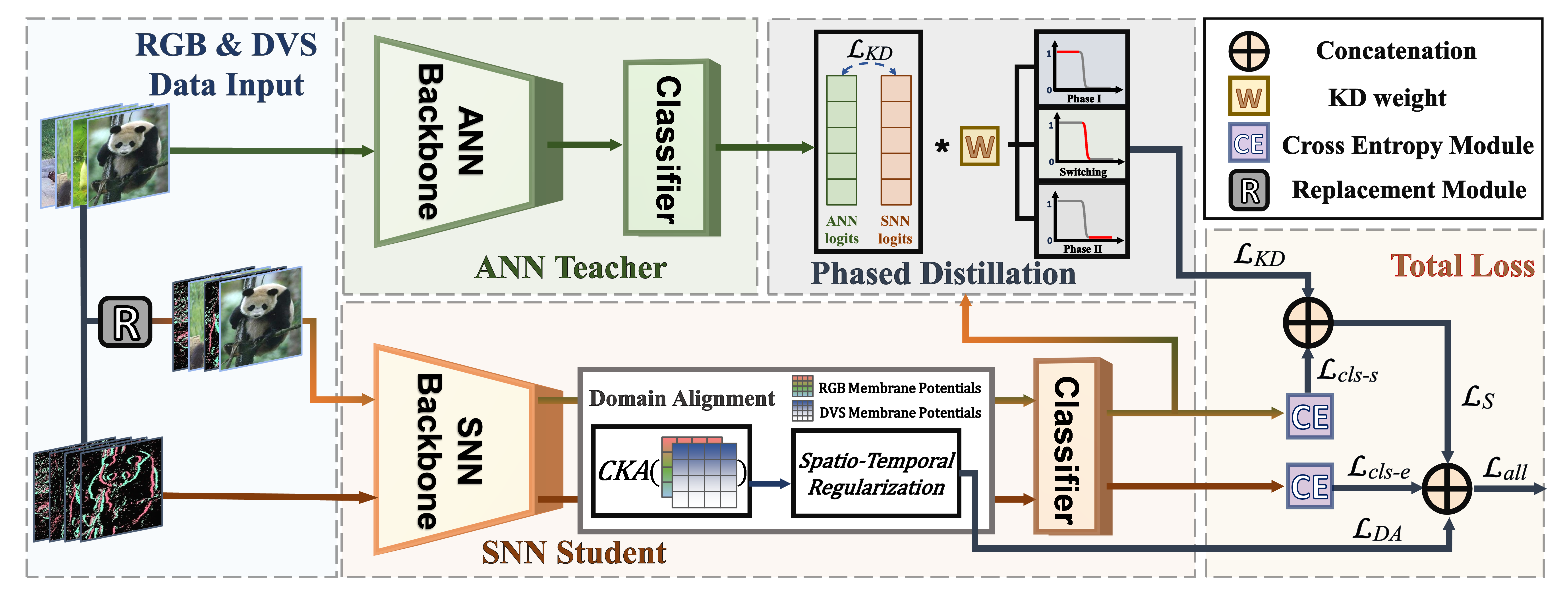}
    \caption{ Overview of our Cross Knowledge Distillation framework between Artificial and Spiking neural networks.
    }
    \label{fig:fig1}
    \vspace{-12pt}
\end{figure*}

\section{Methodology}
\label{sec:method}
In this section, we introduce our cross knowledge distillation (CKD) method, which contains a cross-modality knowledge transfer module and a cross-architecture knowledge distillation module. CKD achieves effective knowledge distillation from ANNs to SNNs under the modality and architecture correspondence. We utilize both RGB images and DVS data during training, but testing is conducted exclusively with DVS data.
\subsection{Cross-Modality Knowledge Transfer}
The cross-modality knowledge transfer module is the basis of our framework, which contains SNN model, domain-alignment module and replacement module. 
\subsubsection{Domain-alignment Module}
Feeding DVS data and its RGB format counterpart into an identical network results in misaligned feature distributions \cite{he:24}, so domain-alignment processing is necessary to safeguard the effectiveness of knowledge transfer.
In this work, we employ a category index to ensure that each pair of inputs corresponds to the same category. Following this, we convert RGB images to HSV (Hue, Saturation, Value) color space and replicate the V channel to match the dimensions of dynamic data in order to feed them in a shared SNN model.

As shown in Fig. \ref{fig:fig1}, domain-alignment loss is employed to constrain the feature distribution differences of the two streams in SNN. 
Here, we use CKA\cite{k:19}, a proven effective index of network representation similarity, to measure that between the two output feature. The closer the value of CKA is to 1, the more correlated the two vectors are. For this reason, we subtract the CKA from 1 as the domain-alignment loss function ${\mathcal{L}_{DA}}$. 
Considering category-alignment and the unique spatial-temporal domain characteristics of SNNs, in particular, the time steps, the domain alignment loss is expressed as:
\begin{equation}\label{eq2}
  {\mathcal{L}_{DA}} = {1 - \frac{1}{T}\sum\limits_{t=1}^T{\underset{y_i = y_j, y\in \mathcal{Y}}{CKA}\left(\bm{F}_s^{i,t},\bm{F}_d^{j,t}\right)}}
\end{equation}
where $T$ denotes the timesteps of input data, $t$ represents current moment, emphasizing that for dynamic data containing rich temporal information, CKA is calculated at each timestep and then averaged across all of them.
The $y$ represents the category of input data, included in the overall set of categories $\mathcal{Y}$. 
The formula $y_i=y_j$ indicates that for paired static and dynamic samples, features $\bm{F}_s^{i,t}$ and $\bm{F}_d^{j,t}$ correspond to inputs from the same category. Moreover, at the temporal level, we add learnable coefficients to assign weight for each timestep.
To prevent the model from overfitting at a certain time step, we adopt DVS data classification loss $\mathcal{L}_{cls\text{-}e}$ as regularization term. In this case, the domain alignment loss is defined as:
\begin{equation}\label{eq3}
  \begin{aligned}
    \mathcal{L}_{DA} &= \frac{1}{T}\sum\limits_{t=1}^T \sigma(\theta_t)(1 - \underset{y_i = y_j, \, y \in \mathcal{Y}}{CKA}(\bm{F}_s^{i,t}, \bm{F}_d^{j,t})) \\
    &+ \frac{1}{T}\sum\limits_{t=1}^T (1-\sigma(\theta_t)) \mathcal{L}_{{cls\text{-}e}}
  \end{aligned}
\end{equation}
where $\sigma$ denotes sigmoid function, $\theta_t$ represents the coefficient at time step $t$.
For classification loss, we employ TET loss which can compensate the momentum loss of surrogate gradient to improve SNN's generalizability \cite{deng:22}.
The overall loss for two streams of static data ${\mathcal{L}_{S}}$ 
can be expressed as:
\begin{equation}\label{eq4}
    \begin{aligned}
    &{\mathcal{L}_{S}} = \alpha {\mathcal{L}_{cls\text{-}s}} + \beta {\mathcal{L}_{DA}}
    \end{aligned}
\end{equation}
where $\alpha $ and $\beta $ are the coefficients of classification and domain-alignment loss. $\mathcal{L}_{cls\text{-}s}$ indicates the classification loss of static data.
\subsubsection{Semantically Similar Replacement Module}
Leveraging the semantic similarity between RGB frame-based data and its event-based (DVS) counterpart, this module replaces static data with DVS data by a non-linear probability function ${P_{replace}}$, which gradually increases from 0 to 1. 
\begin{equation}\label{eq6}
  {P_{replace}} = {\left(\frac{b_i+e_c*b_l}{N_b}\right)^3}
\end{equation}
where $b_i$ is the index within current training batch, $e_c$ is the number of current epoch while $b_l$ is the length of training batches. ${N}_{b}$ represents the sum of batches during the entire training process.

This approach enables DVS data to benefit from the rich knowledge of static data during the early training stage, while progressively shifting the focus to DVS data in later stages, ensuring a smooth and stable transition throughout the process. Moreover, this module builds a more gradual hybrid modal data stream to alleviate the limitation of SNNs in handling large amounts of pure RGB dense data input. This advantage will become more evident in the future, especially in action recognition with more temporal information and video-based detection tasks.

\subsection{Cross-Architecture Knowledge Distillation}

In this work, we use a well-performing ANN teacher model to extract superior knowledge from best-matched modal data, static data, guiding the hybrid data stream within the SNN student model. Subsequently, superior knowledge is transferred from the hybrid stream to the dynamic stream via the cross-modality knowledge transfer module. The distillation between multimodal but same category data effectively bridges the gap of cross-modality and simplifies the problem we face into a more familiar and manageable cross-architecture issue.
\subsubsection{Knowledge Distillation Loss}
\par
\begin{figure}[t]
    \centering
    \begin{minipage}[b]{0.4\linewidth}
        \centering
        \includegraphics[width=\linewidth]{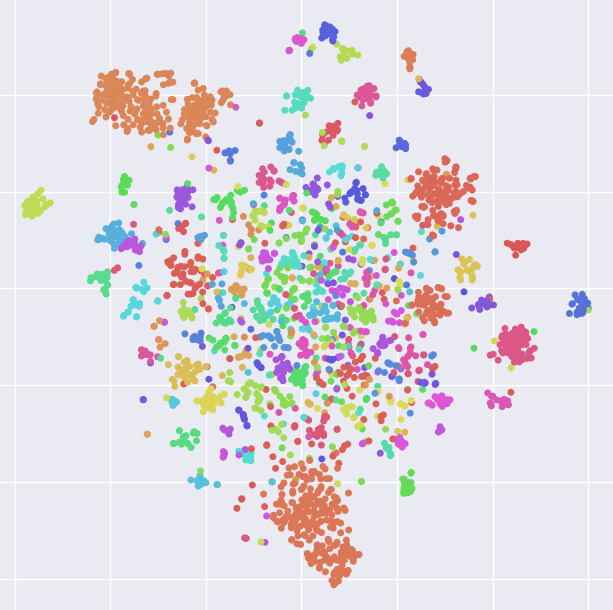}
        \subcaption{}
    \end{minipage}
    \hspace{0.05\linewidth}  
    \begin{minipage}[b]{0.4\linewidth}
        \centering
        \includegraphics[width=\linewidth]{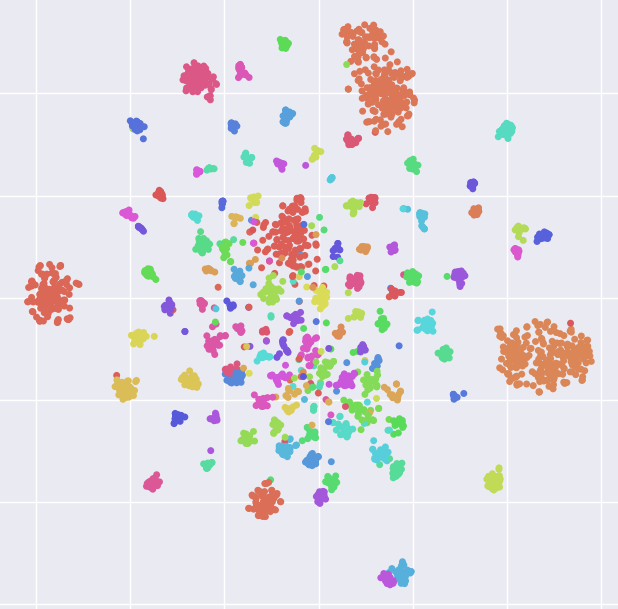}
        \subcaption{}
    \end{minipage}
    \caption{t-SNE visualizations of (a) the baseline and (b) our method on dataset N-Caltech101.}
    \label{fig:fig2}
    \vspace{-15pt}
\end{figure}

 As shown in Fig. \ref{fig:fig1}, we choose logits distillation rather than feature distillation.
On the one hand, our knowledge transfer module already operates in the feature domain, introducing an extra knowledge distillation module is too direct and influential, in contrast, our CKD correct the features indirectly and effectively as shown in Fig.\ref{fig:fig2}. On the other hand, unlike previous cross-architecture feature-domain distillation \cite{deng:21} with strict correspondence between architectural layers, our framework appropriately relaxes the constraints on architecture, thereby reducing the costs of computation and training while expanding the possibilities for cross-architecture distillation. 

In this work, we use vanilla KD loss function to align the logits from the hybrid stream of SNN with those of ANN, distilling valuable information to SNN in logits domain and laying the groundwork for subsequent feature-domain knowledge transfer. The vanilla knowledge distillation loss function $\mathcal{L}_{KD}$, which can be expressed as:
\begin{equation}\label{eq7}
    \mathcal{L}_{KD} = \sum_{i=1}^{N} KL\left( \mathcal{SM}\left(\mathcal{Z}_t^{st}/\tau\right)_i \parallel \mathcal{SM}\left(\mathcal{Z}_s^{st}/\tau\right)_i\right)
\end{equation}
where $\mathcal{Z}_t^{st}$ and $ \mathcal{Z}_s^{st} $ represent the logits of the two static data streams from teacher model and student model respectively. The function $\mathcal{SM}$ denotes the softmax operation, which transforms logits into probability distributions. 
$KL$ denotes the the Kullback-Leibler divergence, a measure of the similarity between two probability distributions, which is computed for each class $i$ here. $N$ denotes the total number of classes while $\tau $ represents the temperature of the distillation.

This indirect knowledge distillation effectively bridges the significant representation gap and mitigates the cross-architecture issue, since the knowledge extracted from the ANN is further corrected by the cross-modality knowledge transfer module, which ensures the mitigation of architectural differences.
\definecolor{lightgray}{HTML}{E3E3E3}

\begin{table*}[htb]
    \centering
    \caption{Experimental results compared with other works. T is the set time steps}
    \label{tab1}
    \vspace{5pt}
    \renewcommand{\arraystretch}{1.2} 
    \begin{tabular}{llllccc} 
        \hline
        \textbf{Dataset} & \textbf{Category} & \textbf{Methods} & \textbf{SNN Model} & \textbf{ANN Model} & \textbf{Timesteps} & \textbf{Accuracy(\%)}\\ 
        \hline
        \multirow{7}{*}{N-Caltech101} & \multirow{2}{*}{Data augmentation}  
                                      & NDA\cite{L:22} & VGGSNN  & -  & 10  & 78.2  \\ 
                                      &               & EventMixer\cite{eventmix}  & ResNet-18  & -  & 10  & 79.2  \\ 
        \cline{2-7} 
                                      & \multirow{3}{*}{Efficient training} 
                                      & TET\cite{deng:22} & VGGSNN & - & 10 & 79.27  \\ 
                                      &                  & TKS\cite{dong:23} & VGGSNN & - & 10 & 84.1    \\ 
                                      &                  & ETC\cite{zhao:23} & VGGSNN & - & 10 & 85.53  \\ 
        \cline{2-7} 
                                      & Domain adaptation                   
                                      & Knowledge-Transfer\cite{he:24} & VGGSNN & - & 10 & $93.18\pm 0.38$ (93.33*)   \\ 
        \cline{2-7} 
                                      & \cellcolor{lightgray}\textbf{Knowledge distillation}  
                                      & \cellcolor{lightgray}\textbf{CKD (Ours)} & \cellcolor{lightgray}\textbf{VGGSNN} & \cellcolor{lightgray}\textbf{WRN101\_2} & \cellcolor{lightgray}10 & \cellcolor{lightgray}\bm{$96.71\pm 0.30$} (\textbf{97.13}) \\ 
        \hline
        \multirow{3}{*}{CEP-DVS}      & Efficient training                   
                                      & TET\cite{deng:22} & ResNet-18 & - & 10 & 25.05  \\ 
                                      & Domain adaptation                   
                                      & Knowledge-Transfer\cite{he:24} & ResNet-18 & - & 10 & $30.05 \pm 0.50$ (35.40*) \\ 
                                      & \cellcolor{lightgray}\textbf{Knowledge distillation}  
                                      & \cellcolor{lightgray}\textbf{CKD (Ours)} & \cellcolor{lightgray}\textbf{ResNet-18} & \cellcolor{lightgray}\textbf{WRN101\_2} & \cellcolor{lightgray}\textbf{10} & \cellcolor{lightgray}\bm{$38.80 \pm 1.23$} (\textbf{40.20}) \\ 
        \hline
        \multicolumn{7}{p{16cm}}{$^{\mathrm{a}}$ \parbox[t]{\linewidth}{The results are mean and standard deviation after taking three different seeds. The symbol ($*$) denotes our implementation of other methods. \\The best accuracy is shown in parentheses. The accuracy of WRN101\_2 is 97.48\% on Caltech101 and 61.45\% on static part of CEP-DVS.}}
    \end{tabular}
    \vspace{-10pt}
\end{table*}

\subsubsection{Phased Distillation Strategy}
As previous outlined, to ensure training stable and smooth, we adopt a semantically similar replacement module. 
Consequently, as training deepens, the hybrid stream which initially designated for static data is gradually incorporated and eventually become dominated by dynamic data.
This phenomenon contradicts our initial intent of using the static stream within the SNN as an intermediary, to ideally mitigate cross-modality and cross-architecture influences.
Therefore, it is crucial to stop the distillation process at a certain point.
The probability of replacement in (\ref{eq6}) 
is rising slowly from 0, even remaining below 0.125 at midpoint.
This level of mixing is acceptable, so we aim to switch the weight of KD from 1 to 0 when KD brings negative impacts.
We use a adjustable function $\gamma(e)$ to enable our switching process manageable:
\begin{equation}\label{eq9}
    {\gamma(e)} = 1 - \frac{1}{1 + \exp(-k \cdot (e - e_{th}))}
\end{equation}
where $e_{th}$ is the threshold of switching epoch, used to control the position of switching, while $k$ is set to control the slope of the switching process.
In this case, the overall loss for two streams of static data $\mathcal{L}_{S}$ in (\ref{eq4}) could be redefined as:
\begin{equation}\label{eq8}
    \begin{aligned}
    &{\mathcal{L}_{S}} = \alpha {\mathcal{L}_{cls-s}} + \beta {\mathcal{L}_{DA}} + \gamma(e) {\mathcal{L}_{KD}} 
    \end{aligned}
\end{equation}
where the ${\mathcal{L}_{KD}}$ is denied as shown in (\ref{eq7}). 
The total training loss can be expressed as $\mathcal{L}_{all} = \mathcal{L}_{S} + \mathcal{L}_{cls\text{-}e}$.

\section{Experiments}
We conduct experiments on mainstream event-based datasets N-Caltech101 \cite{Ncaltech} with its static version Caltech101 \cite{Caltech101}, and the image-event paired dataset CEP-DVS \cite{deng:21}, to evaluate the effectiveness of the proposed method.

\subsection{Implement Details}
We integrate all DVS data into frames and resize them to 48x48 for both the N-Caltech101 and CEP-DVS datasets. For a fair comparison, we use the VGGSNN model \cite{he:24} trained on N-Caltech101 for 300 epochs with 10 time steps, and the Spiking-ResNet18 model \cite{he:24} trained on CEP-DVS for 200 epochs with 6 time steps. We select the Wide\_ResNet101\_2 (WRN101\_2) \cite{spinal} as ANN teacher model, which is finetuned 20 epochs from ImageNet in HSV-converted Caltech101. Additionally, we set both the $\alpha$ and $\beta$ in (\ref{eq8}) to 1. All experiments are conducted using the BrainCog framework \cite{zeng:23}.

\begin{table}[t]
    \renewcommand\arraystretch{1.2} 
    \centering    
    \caption{Ablation studies of different knowledge distillation loss functions on CKD framework.} 
    \label{tab2}
    \begin{tabular}{llc|llc} 
        \hline
        \textbf{Network} & \textbf{Methods} & \textbf{Acc(\%)} & \textbf{Network} & \textbf{Methods} & \textbf{Acc(\%)} \\ \hline
        \multicolumn{3}{c|}{N-Caltech101} & \multicolumn{3}{c}{CEP-DVS} \\ \hline
        \multirow{5}{*}{VGGSNN} & baseline & 93.18 & \multirow{5}{*}{ResNet-18} & baseline & 30.50 \\ \cline{2-3} \cline{5-6}
        & DKD      & 96.48 &  & DKD      & 37.63 \\ 
        & LumiNet  & 95.82 &  & LumiNet  & 38.43 \\ 
        & DIST     & 95.59 &  & DIST     & 37.75 \\ 
        & \textbf{KD}       & \textbf{96.71} &  & \textbf{KD}       & \textbf{38.80} \\ \hline
        \multicolumn{6}{p{7cm}}{$^{\mathrm{a}}$ \parbox[t]{\linewidth}{The results are mean after taking three different seeds.}}
    \end{tabular}
    \vspace{-15pt}
\end{table}

\subsection{Comparison with the State-of-the-Art}
We first evaluate our CKD on N-Caltech101 and compare it with TET \cite{deng:22}, TKS \cite{dong:23}, ETC \cite{zhao:23}, and Knowledge-Transfer \cite{he:24}. The results presented in TABLE \ref{tab1} demonstrate that our method outperforms all the compared methods. We report a new state-of-the-art accuracy of 97.13\% on N-Caltech101, which is close to that of our ANN teacher model \cite{spinal} (97.48\%) and the current state-of-the-art for Caltech101 \cite{SotaCaltech101} (98.02\%).

The results show that with our CKD, SNNs can perform almost as well as ANNs on image classification tasks. At the same time, our CKD achieves cross-modal and cross-architecture knowledge distillation. It also ensures modality-architecture correspondence, thereby fully leveraging the advantages of SNNs, such as energy efficiency, biological plausibility, and the asynchronous nature and high dynamic range of DVS data. This lays a solid foundation for future deployment on brain-inspired chips.
As shown in the t-SNE visualization in Fig. \ref{fig:fig2}, compared to our baseline \cite{he:24}, the data points corresponding to different classes are distinctly more separated, with tighter clusters and less overlap between classes. This indicates that knowledge extracted from the ANN has been efficiently transferred to the SNN, leading to more discriminative features and higher accuracy.

\subsection{Ablation Studies}
To verify the effect of CKD, we conduct ablation studies on KD loss function and the phase switching function.
As shown in TABLE \ref{tab2}, on the N-Caltech101, we achieve a 97.13\% accuracy using vanilla KD. In contrast, DKD \cite{DKD}, LumiNet \cite{luminet} and DIST \cite{DIST} yield relatively lower accuracies than vanilla KD, since these methods exert relatively excessive influence on the static stream within the SNN, adversely affecting our domain-alignment module and leading to a slightly decline in accuracy. However, they still surpass our baseline by over 2\%, demonstrating the superiority of our CKD framework. On the CEP-DVS, the results are similar, further validating our conclusion and the effectiveness of CKD.

\begin{figure}[t]
    \centering
    \begin{minipage}[b]{0.45\linewidth}
        \centering
        \includegraphics[width=\linewidth]{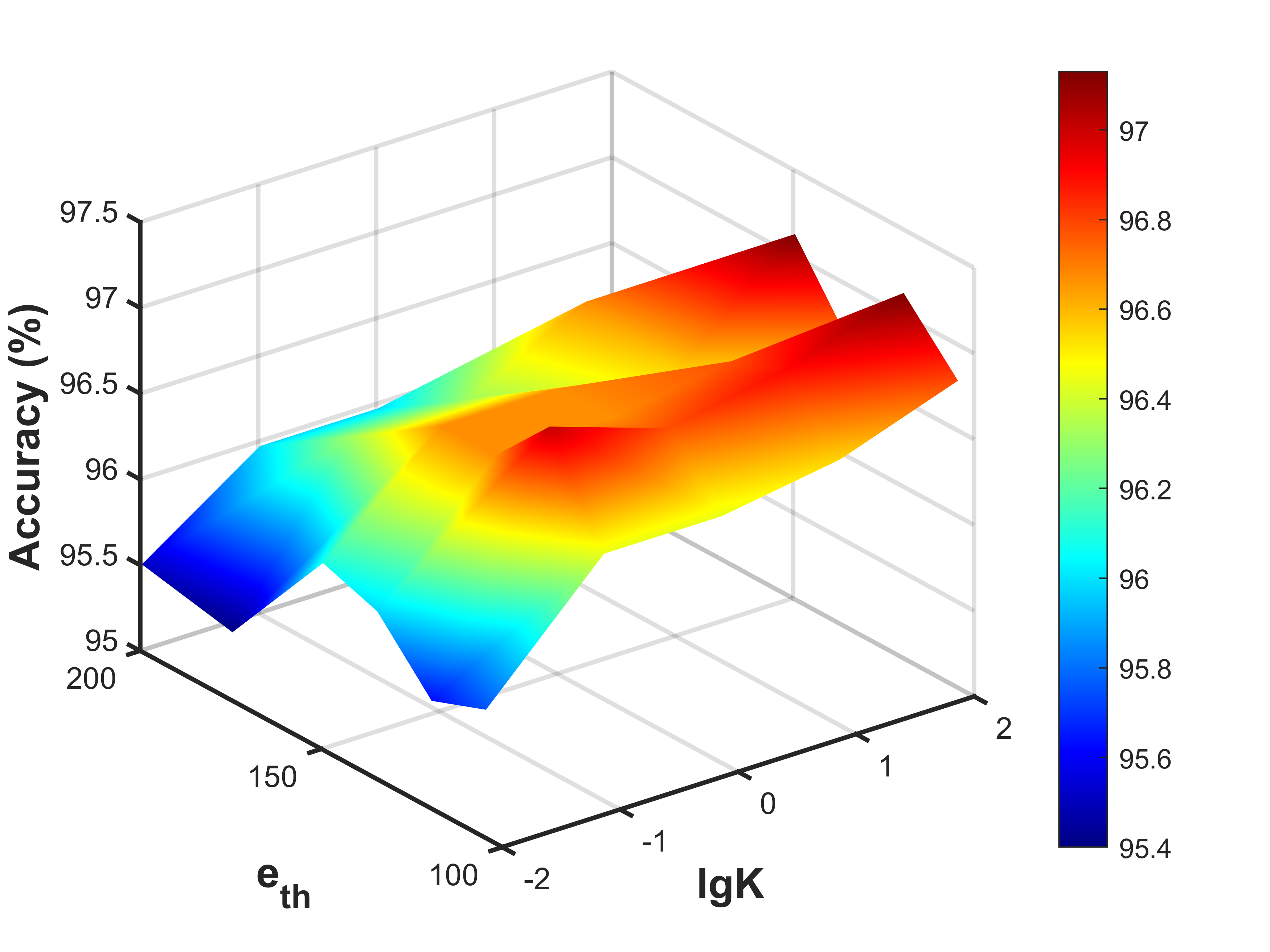}
        \subcaption{}
    \end{minipage}
    \hspace{0.05\linewidth}  
    \begin{minipage}[b]{0.45\linewidth}
        \centering
        \includegraphics[width=\linewidth]{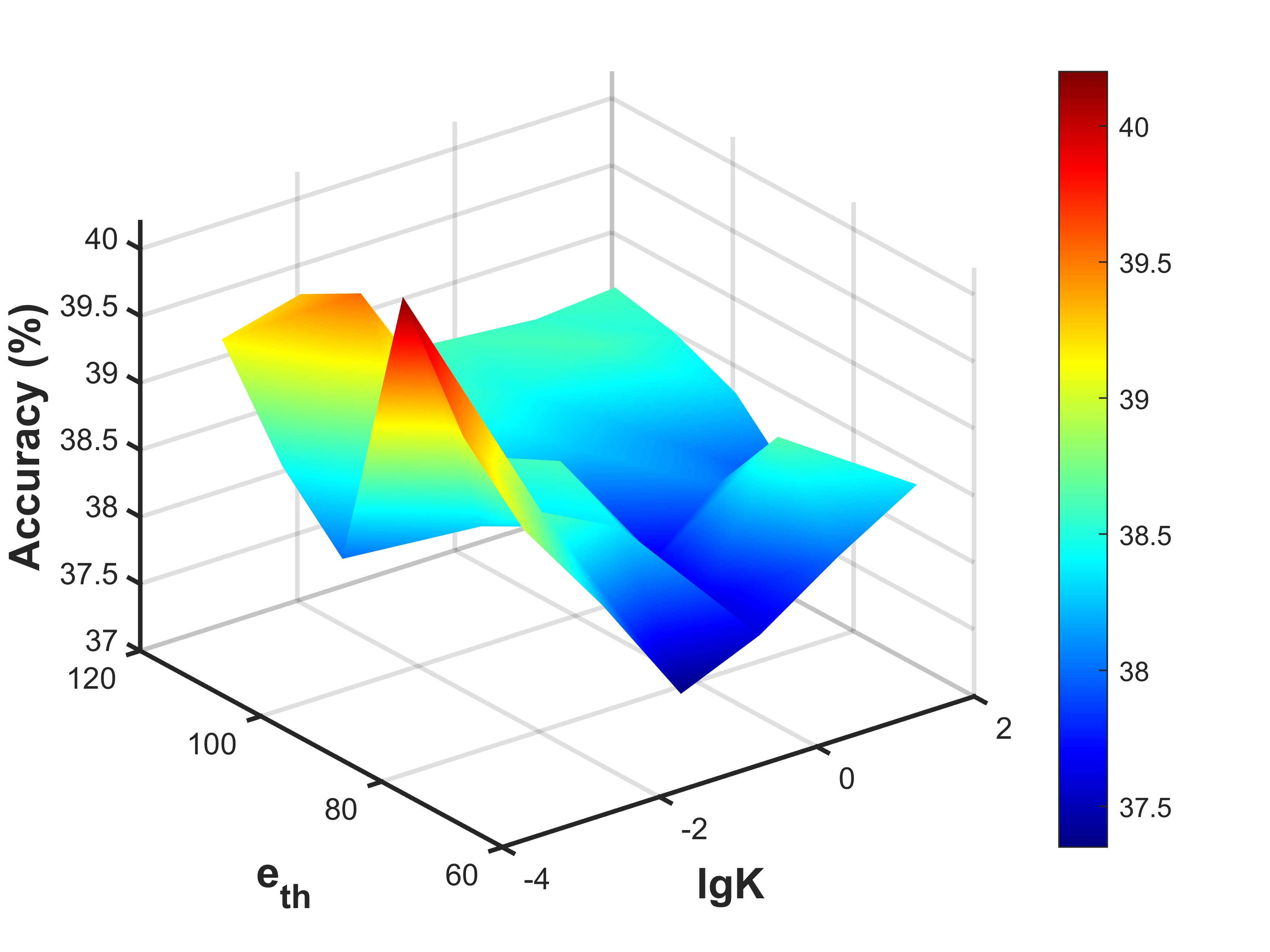}
        \subcaption{}
    \end{minipage}
    \caption{Visualization of phase switching function parameters ablation study on (a) N-Caltech101 and (b) CEP-DVS.}
    \label{fig:fig3}
    \vspace{-15pt}
\end{figure}

For the other ablation study, we perform visualizations using various sets of parameters $e_{th}$ and $k$ in phase switching function. As shown in Eq. \ref{eq9}, the $k$ is set to control the slope of our switching function, when $k$ becomes greater, the switching becomes swifter. The $e_{th}$ controls the beginning epoch of the switching. As shown in Fig. \ref{fig:fig3}, we apply base-10 logarithm to the $k$ values to improve their readability. 

On the N-Caltech101, peaks (97.13\%) occur at $e_{th} = 149.5, k = 100$ and $e_{th} = 119.5, k = 100$ while declining when $k$ turns small or $e_{th}$ deviates excessively from midpoint. The lowest point (95.40\%) occurs at $e_{th} = 174.5, k = 0.01$. These results indicate that on this dataset the performance of this SNN model improves the most when the switching is swift and occurs around the midpoint, whereas the effect is relatively modest during a delayed and smooth switching process. On the CEP-DVS, peak (40.2\%) are observed at small values of $k$, with $e_{th} = 89.5, k = 0.001$, representing that slower and steadier phase switching allows SNNs to learn more and better from ANNs. The lowest point (37.35\%) occurs at $e_{th} = 69.5, k = 0.1$. These visualizations illustrated the effects of different switching strategies. Nevertheless, all results from this ablation study surpass our baseline, validating the effectiveness of our CKD method.

\section{conclusion}
In this paper, we proposed cross knowledge distillation (CKD), a novel method to bridge both architectural and modal differences between ANNs and SNNs as well as RGB and DVS data. Through extensive experiments on main-stream neuromorphic datasets, we proved the effectiveness of CKD. We achieve a new SOTA top-1 accuracy of 97.13\% on N-Caltech101, a significant improvement over previous works, as well as a competitive accuracy of 40.20\% on CEP-DVS. Our results show that SNNs, with the help of CKD, can perform on par with ANNs, making them viable for real-world vision tasks. Moreover, the proposed method lays a solid foundation for future research, enabling SNNs to tackle more complex and temporally demanding tasks, potentially paving the way for their deployment in neuromorphic computing systems.

\section{Acknowledgments}
 This work was supported by the Ningbo Municipal Natural Science Foundation of China (No. 2022J114), National Natural Science Foundation of China (No. 62271274), Ningbo S\&T Project (No.2024Z004) and Ningbo Major Research and Development Plan Project (No.2023Z225)

\bibliographystyle{IEEEbib}
\bibliography{icme2025references}

\vspace{12pt}

\end{document}